\documentclass[runningheads]{llncs}

\usepackage{eccv}

\usepackage{eccvabbrv}

\usepackage{graphicx}
\usepackage{adjustbox}
\usepackage{booktabs} 
\usepackage{algorithm}
\usepackage{algpseudocode}
\newcommand{\ua}{\uparrow}
\newcommand{\da}{\downarrow}
\newcommand{\shortcite}{\cite}

\usepackage[accsupp]{axessibility}  

\usepackage{hyperref}

\usepackage{orcidlink}

\usepackage{graphics}
\usepackage{array}
\usepackage{multirow}
\usepackage{wrapfig}
\usepackage{hhline}
\usepackage{pifont}
\newcommand{\showimagew}[2][\linewidth]{\includegraphics[width={#1}]{{#2}}}

\DeclareMathOperator*{\argmax}{arg\,max} 
\newcommand{\reffig}[1]{Figure~\ref{#1}} 
\newcommand{\topscore}[1]{\textcolor{blue}{\textbf{#1}}} 
\newcommand{\udl}{\underline} 
\newcolumntype{K}[1]{>{\centering\arraybackslash}p{#1}}

\usepackage{atbegshi}
\newcommand{\placetextbox}[3]{
  \setbox0=\hbox{#3}
  \AtBeginShipoutNext{\AtBeginShipoutUpperLeft{%
    \put(\dimexpr#1\paperwidth\relax,-\dimexpr#2\paperheight\relax)
    {\vtop{{\null}\makebox[0pt][c]{#3}}}%
  }}%
}

\newcommand{\mcal}[1]{\mathcal{#1}}

\begin{document}

\title{A Bottom-Up Approach to Class-Agnostic \\Image Segmentation}

\titlerunning{A Bottom-Up Approach to Class-Agnostic Image Segmentation}

\author{
Sebastian Dille\inst{1}\orcidlink{0000-0003-0390-2803}
\and 
Ari Blondal\inst{1,2}\orcidlink{0009-0001-9807-7385}
\and
Sylvain Paris\inst{3}
\and
Ya\u{g}{\i}z Aksoy\inst{1}\orcidlink{0000-0002-1495-0491}
}

\authorrunning{S.~Dille et al.}

\institute{
Simon Fraser University, BC, Canada \and
McGill University, QC, Canada \and
Adobe Research, MA, United States
}

\maketitle

\placetextbox{0.14}{0.03}{\includegraphics[width=4cm]{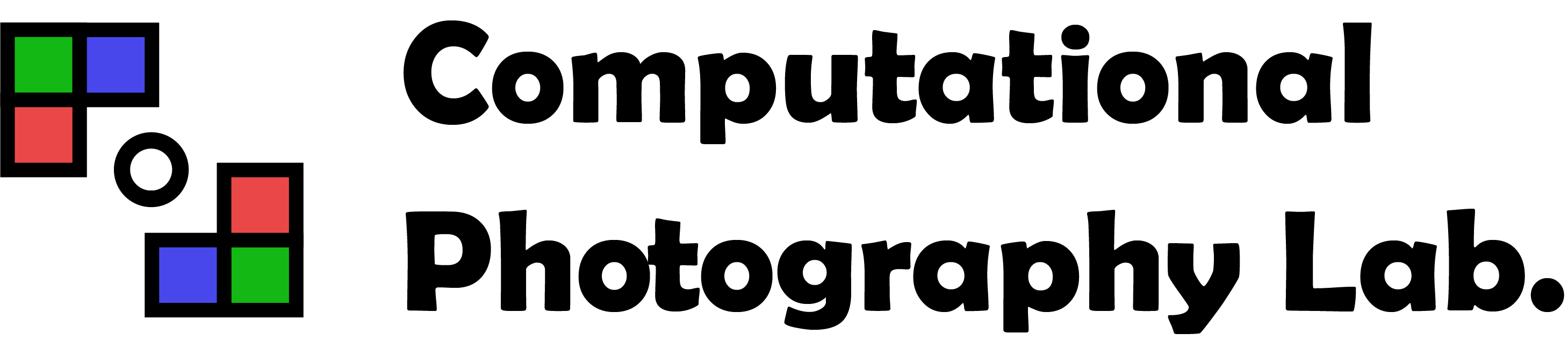}}
\placetextbox{0.8}{0.03}{Find the project web page here:}
\placetextbox{0.8}{0.045}{\textcolor{purple}{\url{https://yaksoy.github.io/bottomup/}}}

\begin{center}
    \centering
    \showimagew[\linewidth]{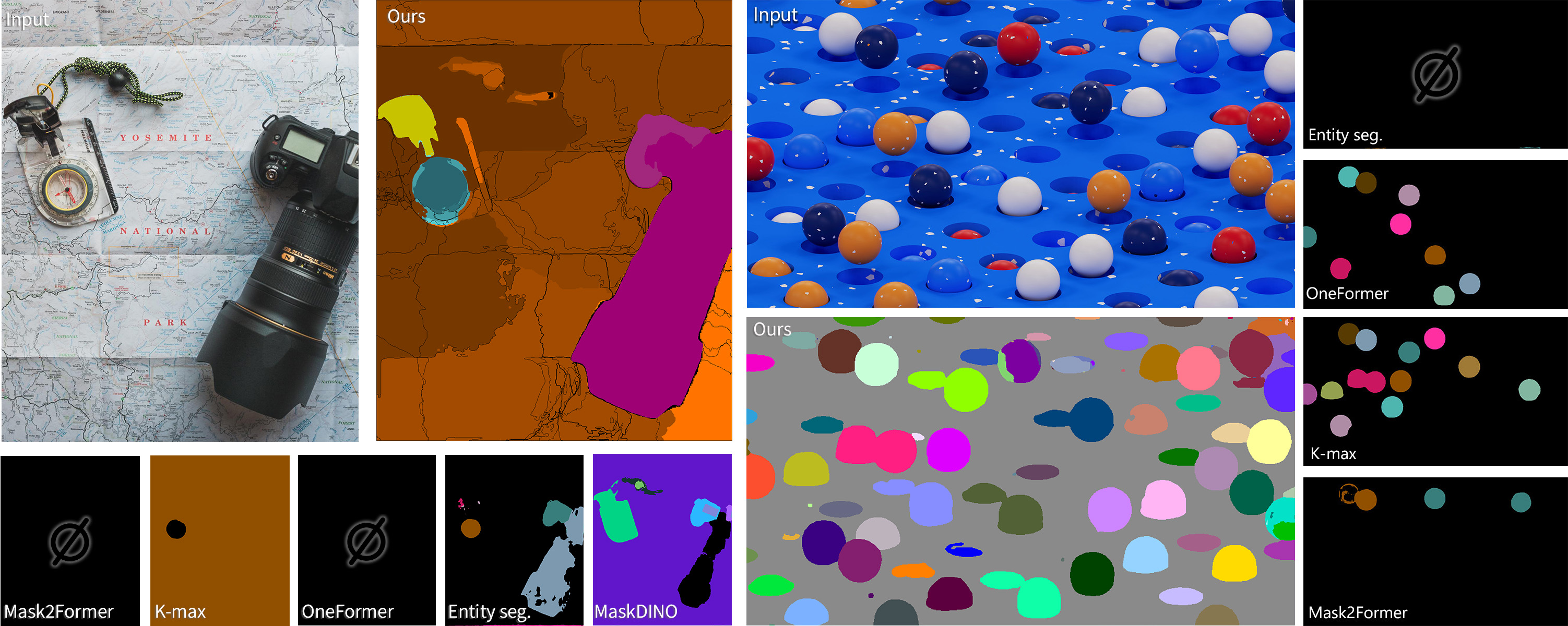}
    \captionof{figure}{
      We introduce a bottom-up approach to class-agnostic image segmentation. We show that our formulation leads to generalization to images in-the-wild that are not well-represented in common training datasets. We generate detailed segmentation maps for complex scenes where other class-based or class-agnostic approaches fall short.
    }
    \label{fig:teaser}
\end{center}%

\begin{abstract}
Class-agnostic image segmentation is a crucial component in automating image editing workflows, especially in contexts where object selection traditionally involves interactive tools. 
Existing methods in the literature often adhere to top-down formulations, following the paradigm of class-based approaches, where object detection precedes per-object segmentation.
In this work, we present a novel bottom-up formulation for addressing the class-agnostic segmentation problem. 
We supervise our network directly on the projective sphere of its feature space, employing losses inspired by metric learning literature as well as losses defined in a novel segmentation-space representation.
The segmentation results are obtained through a straightforward mean-shift clustering of the estimated features.
Our bottom-up formulation exhibits exceptional generalization capability, even when trained on datasets designed for class-based segmentation. We further showcase the effectiveness of our generic approach by addressing the challenging task of cell and nucleus segmentation.
We believe that our bottom-up formulation will offer valuable insights into diverse segmentation challenges in the literature.
\end{abstract}
\section{Introduction}
\label{sec:intro}

In most image editing scenarios, object selection is the first step for localized image editing or compositing. 
Automating the object selection, hence, is an interesting application scenario for increased productivity. 
A generalized object selection method requires the segmentation of every object in any image in the wild. 
This is a challenging task that standard class-based image segmentation approaches such as semantic or panoptic segmentation fail to accomplish due to the inherently limited number of classes labeled in a training dataset.

With this motivation, recent literature focuses on the \emph{class-agnostic} segmentation problem. 
Open-set panoptic segmentation \cite{hwang2021exemplar,xu2022dual} approaches this problem by extending the label space in panoptic segmentation with an \emph{unknown} class, aiming to detect objects that do not fit in the set of defined classes. 
Open-world entity segmentation \cite{qiOpenWorldEntitySegmentation2021,qilu2022finegrained}, on the other hand, defines the segmentation problem as fully class-agnostic, and presents a method that can detect object centers, which are then used to create the final segmentation map. 
Segment anything \cite{kirillov2023segment} adopts a prompt-based approach and conducts class-agnostic segmentation by assuming a regular grid of user inputs, powered by an extensive dataset. 

All these approaches follow a \emph{top-down} approach to segmentation, where the first task of the system is to detect the objects in the scene, followed by per-object segmentation.
This top-down approach is in contrast with our understanding of human cognition. 
Humans can easily identify objects or coherent regions in a wide variety of realistic or abstract, complex or simple images. 
The dominant process in human object detection is modeled to be \emph{bottom-up} \cite{humanUnderstanding}, where grouping of features in the scene is followed by object detection and finally classification.

In this work, we present a novel approach to class-agnostic image segmentation with a bottom-up formulation.
We adopt the entity definition by Qi~\etal~\cite{qiOpenWorldEntitySegmentation2021} that unifies \emph{things} and \emph{stuff} into classless entities. 
We develop our formulation in a feature space with projective geometry, generating per-pixel features that are parallel to each other within the same entity, and orthogonal to all features outside their entity.
This allows for maximally separated entities that can conveniently be clustered with simple mean-shift clustering for a dense class-agnostic segmentation during inference.
We achieve this with a loss combination inspired by metric learning and a novel segmentation-space formulation that allows for the backpropagation of segmentation-focused losses into our hyper-dimensional feature space.

Our formulation is carefully developed for generalization to class-agnostic only through class-based datasets.
Despite using the standard segmentation datasets MS COCO \cite{lin2014microsoft,caesar2018coco}, ADE-20k \cite{zhou2018ade20k}, and CIHP \cite{cihp} as our only real-world training data, our bottom-up approach shows an exceptional generalization ability to unseen classes as well as out-of-distribution images as Figure~\ref{fig:teaser} shows.
We demonstrate the performance of our system through zero-shot quantitative analysis.
Despite utilizing a smaller architecture, we show that we can generate detailed segmentations for complex scenes in the wild.  
We further demonstrate the generic nature of our bottom-up formulation by improving upon the state-of-the-art in cell and nucleus segmentation.

\section{Related Work}
\label{sec:related}

The field of automatic image segmentation is dominated by class-based object labeling approaches~\cite{long2015fully,Ronneberger2015UNetCN,chen2017deeplab,He2020MaskR,fathiSemanticInstanceSegmentation2017, cascadercnn, wang2021solo, wang2020solov2, tian2020conditional,kirillov2019panoptic, panopticfpn, cheng2020panoptic, li2022panoptic, cheng2021maskformer, cheng2022masked}. These methods are trained to recognize objects from a fixed set of known classes and assign pixel labels accordingly. Depending on the application scenario, the algorithms are either identifying semantics alone~\cite{long2015fully,Ronneberger2015UNetCN,chen2017deeplab}, distinguishing individual instances of countable objects~\cite{He2020MaskR,fathiSemanticInstanceSegmentation2017, cascadercnn, wang2021solo, wang2020solov2, tian2020conditional}, or combining both in a panoptic fashion~\cite{kirillov2019panoptic, panopticfpn, cheng2020panoptic, li2022panoptic, cheng2021maskformer, cheng2022masked}. Their inherent inability to generalize to unseen classes, however, makes them less suitable for use in image editing. 
We focus on class-agnostic segmentation below and refer to the recent survey \shortcite{minaee2021image} for an in-depth review.

\vspace{2mm}\noindent\emph{Class-agnostic Image Segmentation} \quad
Recently, a growing number of segmentation approaches~\cite{qiOpenWorldEntitySegmentation2021,qi2022ssl,qilu2022finegrained,hwang2021exemplar,xu2022dual} are removing class dependency to handle out-of-distribution objects and improving generalization: 
Open-set panoptic segmentation methods~\cite{hwang2021exemplar,xu2022dual} on the one hand are still closely following the concept of panoptic segmentation but introduce an additional class to the training set to label unknown elements. Once identified, the corresponding areas are further segmented via class-agnostic clustering based on predicted bounding boxes.

Entity-segmentation methods~\cite{qiOpenWorldEntitySegmentation2021,qi2022ssl,qilu2022finegrained} on the other hand entirely remove semantic information from the training process, treating each object in the dataset as a unique entity. Qi~\etal\shortcite{qiOpenWorldEntitySegmentation2021} first introduce this concept by replacing the supervision from a proposal-based segmentation approach~\cite{tian2020conditional} with class-agnostic masks. 
They show how this change alone results in increased generalization capability and further adapt the concept in subsequent work for pretraining in class-based segmentation~\cite{qi2022ssl} and to generate high-resolution results~\cite{qilu2022finegrained}.
Both works formulate segmentation in a top-down fashion, incorporating a proposal generator to predict bounding boxes \cite{qi2022ssl} or entity centers \cite{qilu2022finegrained}. 
This limits their generalization ability to objects that do not match the training distribution in appearance. 
In contrast, we construct our method as a bottom-up framework based on \emph{object discrimination} that generates segments by clustering on the hypersphere and is independent of the exact appearance. 

Segment anything \cite{kirillov2023segment} takes inspiration from prompt-based natural language processing approaches and formulates the segmentation problem with various forms of input. 
For the class-agnostic problem akin to entity segmentation, they assume a regular grid of input prompts to generate their dense output and demonstrate a strong generalization ability enabled by their immense dataset. 
One major shortcoming of prompt-based methods is the dependence on very large training datasets, which limits their applicability to other domains where collection of such datasets is prohibitively expensive such as commercial applications, medical image segmentation, and fine-grained segmentation. 

Our bottom-up approach, on the other hand, is designed to leverage small or incomplete datasets while still achieving generalizability. 
We achieve similar performance to the segment anything model with similar number of parameters to ours, despite them training on a dataset that is larger than ours by 2 orders of magnitude.
We also achieve state-of-the-art performance in cell and nucleus segmentation compared to domain-specific approaches, demonstrating the generic use of our bottom-up formulation in problems with limited training data. 
Our representation-based metric learning approach can further be integrated into future prompt-based approaches to improve their performance in problems where large-scale data acquisition has inherent challenges.

\vspace{2mm}\noindent\emph{Entity Representation in Projective Space} \quad
Object discrimination is a long-existing concept in image processing in the form of clustering-based methods~\cite{nock2004statistical,yi2004automated,ji1998image,tremeau1997region,luo1998incorporation,cheng2000hierarchical} and is recently being revisited by instance segmentation approaches with the goal to distinguish individual instances within an already recognized ``thing''-category~\cite{gao2019ssap,de2017semantic,uhrig2018box2pix, Bai_2017_CVPR,kong2018recurrent,cheng2020panoptic}. In this constrained setting, the prediction accuracy can be greatly increased by computing affinities on high-dimensional features instead of pixels and applying contrastive losses as class-agnostic supervision. For panoptic segmentation, the bottom-up approach is challenging since the combination of different elements within the same ``stuff'' ground truth category creates ambiguities,
and so far only combined approaches have been proposed with bottom-up ``thing'' discrimination and top-down segmentation for ``stuff''~\cite{cheng2020panoptic}. We argue that a careful definition of the feature representation and supervision space is crucial for bottom-up panoptic segmentation.

We formulate our supervision thus on the projective sphere, a hypersphere with antipodal equivalence.
Non-euclidean representations have seen growing attention in grouping tasks due to the inherent hierarchical properties in hyperbolic space~\cite{nickel2017poincare,khrulkov2020hyperbolic,weng2021unsupervised,atigh2022hyperbolic, cetin2023hyperbolic,ge2023hyperbolic,lin2023mhcn} and intuitive metric learning on the hypersphere~\cite{kong2018recurrent, Hwang_2019_ICCV,chen2020simple,he2020momentum,zhou2022rethinking}.
In line with works by Kong~\etal~\cite{kong2018recurrent} and Hwang~\etal~\cite{Hwang_2019_ICCV}, we use cosine similarity to define a metric on the projective sphere.
By supervising directly in this feature space, we yield an entity-specific representation that allows us to ignore ambiguous background regions during training and to apply simple mean-shift clustering during inference.

\section{Generating Distinguishable Features}
\label{sec:features}

We approach the class-agnostic image segmentation problem by looking back at the most basic definition of image segmentation.
Our aim is to generate an image representation that allows us to cluster the pixels in the image into segments that correspond to different entities. 
For this purpose, we formulate our training and inference scheme purely in our projective spherical feature space using losses inspired by contrastive learning. 
We also define a low-dimensional segmentation space that allows us to signal clustering performance to the network during training. 
We show that our features can be used to segment the image using a simple mean-shift clustering formulation.

\subsection{Feature Representation}
We define our feature space as the real projective sphere. This means that it resembles a hyper-dimensional sphere of radius 1, where each feature - encoded through unit homogeneous coordinates - forms a point on the surface, and points on opposite sides of the sphere are equivalent. Because of the equivalency, these points resemble \emph{lines} through the origin in $d$-dimensional space, a concept that also better illustrates our representation's focus on angle distance, parallelism, and orthogonality.
Our feature space is thus defined as $\mathcal{F}$:
\begin{equation}
\label{eq:method:featurespace}
    \mathcal{F} = \{\vec{f} \in \mathbb{R}^d : ||\vec{f}||_2 = 1, \vec{f} = \vec{-f} \},
\end{equation}
and we endow it with a distance metric defined as the cosine distance between two lines:
\begin{equation}
    dist(\vec{f}_1, \vec{f}_2) = 1 - |\vec{f}_1 \cdot \vec{f}_2|.
\end{equation}
Given this setup, we want features corresponding to pixels from the same object to be parallel, and features from different objects to be orthogonal.
This is a powerful formulation with a continuous and piece-wise differentiable distance function between any two features.
It allows for up to $d$ objects with features maximally far apart.
This is in contrast to setups that use directions \cite{kong2018recurrent} or bounded points \cite{Bai_2017_CVPR} as features rather than lines, where there is only a single maximally distant feature to any other.
The desired orthogonality of features from different segments plays a crucial role in defining our low-dimensional segment-space representation and in handling training data with incomplete labels, as we discuss later in this section.
The output of our network is defined as a $h \times w \times d$ dimensional feature map, $h$ and $w$ being the height and width of the input image and we set $d$ to be 128. 
We normalize the estimated features to unit length to get our per-pixel features $\vec{f} \in \mathcal{F}$.
During training, we also define an L2 regularization loss that signals our network to generate unit-length features:
\begin{equation}
    \mathcal{L}_u = \sum_i |1 - \|\vec{f}_i\|_2|.
\end{equation}

\subsection{Determining Target Lines for each Entity}
We aim to generate a feature for every pixel such that the feature of a pixel is parallel to others that belong in the same entity, and orthogonal to the ones that belong in others. 
This goal is shared with many standard metric or affinity learning formulations \cite{banerjee2005clustering, gopal2014mises, kong2018recurrent}. 
However, defining a loss function on all the inter-pixel affinities quickly becomes prohibitively expensive due to the quadratic explosion of the $N\times N$ possible pixel pairings, $N=h\times w$.

Instead, we first determine target lines for each entity during training and define losses that align each pixel's feature with its corresponding entity, while pushing it away from all other target lines. This simplifies our optimization problem from a many-to-many comparison setting to many-to-few. We will also use target lines to define our segmentation-space as described later in this section.

For each entity available in the ground-truth, we calculate the target line using the predicted features of all the pixels that belong to that entity.
For homogeneous coordinates, euclidean averaging of the features may result in a degenerate average. Instead, we compute the average orientation $\vec{\mu}_k$ through
\begin{equation}
\label{eq:method:quaternionAvg}
    \vec{\mu}_k = \argmax_{\vec{v}}{\vec{v}^T M_k \vec{v}},
    \quad
    M_k = \sum_{i \in \mathcal{E}_k} \vec{f}_i \vec{f}^T_i,
\end{equation}
where $\mathcal{E}_k$ is the set of pixels belonging to the ground-truth entity $k$.
The solution of this maximization problem is given by the eigenvector of $M_k$ corresponding to its largest eigenvalue. In line with our feature space definition in Eq.~\ref{eq:method:featurespace}, $\vec{\mu}_k$ is of unit-length and changing its sign, or the sign of any $\vec{f}_i$ does not affect the result. 
Markley~\etal~\cite{markley2007averaging} presents a comprehensive exposition of this orientation averaging approach in the case of quaternions.

\subsection{Handling Imperfect Ground-truth}
\label{sec:method:bg}
Most large-scale datasets with annotated ground-truth segmentation have been collected for class-based segmentation approaches such as semantic or panoptic segmentation. 
Due to the inherently limited set of classes a dataset contains, many objects that are not in one of the pre-defined classes are not segmented but either included in a general \emph{background} category or just lack any label. 
Due to the complexity of annotating every single object in an image, even the class-agnostic SA-1B Dataset \cite{kirillov2023segment} contains many unlabeled objects. 

As the goal of our class-agnostic segmentation approach is to generalize to any object, we can not treat the background category as its own entity. There may be multiple objects in the background category and without individual ground-truth labels, we can not determine a target line to align all the pixels. 
However, as an ideal representation in our feature space has orthogonal lines for each entity, we know that we want all the features in the background category to be orthogonal to the target lines $\vec{\mu}_k$ for all known entities in the image. 
Hence, when formulating our loss functions, we will exclude the background category for all losses promoting alignment and include them in ones that promote orthogonality.
This way, we promote features in the background that are distinguishable from the known entities while not punishing the network for correctly estimating entities that are not annotated in the ground-truth.

Due to the complexity of annotating many segments with pixel-perfect precision, segmentation datasets often have boundary inaccuracies in the ground-truth labels. 
As a result, including the possibly inaccurate boundaries in the loss formulation harms the boundary accuracy of the system. 
To address this, we erode all the ground-truth annotations with a $5\times5$ kernel and exclude the eroded-away pixels from all loss computations.

\subsection{Attraction and Repulsion}
\label{sec:metriclosses}

To align the features of the pixels that belong in a single entity to their target line $\vec{\mu}$, we define a simple \emph{attraction} loss:
\begin{equation}
    \mathcal{L}_a = \frac{1}{K} \sum_k \frac{1}{|\mathcal{E}_k|} \sum_{i\in\mathcal{E}_k} \left( 1 - |\vec{f}_i \cdot \vec{\mu}_k | \right),
\end{equation}
where $K$ is the number of labeled entities in the image. 
Rather than pairwise aligning all features, this simple loss pulls every pixel in an entity towards alignment with the same target line. 

Similarly, we define a \emph{repulsion} loss that pushes every pixel to be orthogonal to the target lines of other entities:
\begin{equation}
    \mathcal{L}_r = \frac{1}{\sqrt{K+1}} \left( \mathcal{L}_{r}^{BG} + \sum_k \frac{1}{|\mathcal{E}_k|} \sum_{l \neq k} \sum_{i \in \mathcal{E}_k} | \vec{f}_i \cdot \vec{\mu}_l | \right)
\end{equation}

As noted in Sec.~\ref{sec:method:bg}, using an average orientation $\vec{\mu}_{BG}$ of the background features, we include the background pixels in the repulsion loss:
\begin{equation*}
    \mathcal{L}_{r}^{BG} = \sum_k \frac{1}{|\mathcal{E}_k|} \sum_{i \in \mathcal{E}_k} | \vec{f}_i \cdot \vec{\mu}_{BG} | +  \frac{1}{|BG|} \sum_{i \in BG} \sum_k |\vec{f}_i \cdot \vec{\mu}_k|,
\end{equation*}
that pushes the pixels in the background category to be orthogonal to all entity target lines, as well as the features in known entities to be orthogonal to the primary background orientation. We found that normalizing the repulsion loss with $K+1$ makes it ineffective for images with many segments. Instead, we use $\sqrt{K+1}$, providing a good balance for images with few or many entities. 

During training, our network may fail to differentiate between two different entities and generate similar features for both. In such a case with $\vec{\mu}_k \approx \vec{\mu}_l$, the attraction and repulsion losses cancel each other. 
This results in a lack of a loss that signals the network to separate the two entities from each other. 
To promote separation between entities, we add the sparse regional contrast
loss $\mcal{L}_{rc}$ introduced by Liu~\etal~\cite{liu2022bootstrapping} as a second contrastive supervision, empirically setting the temperature $\tau = 0.5$ and using $256$ queries per entity.

\begin{figure}[t]
  \centering
  \showimagew[\textwidth]{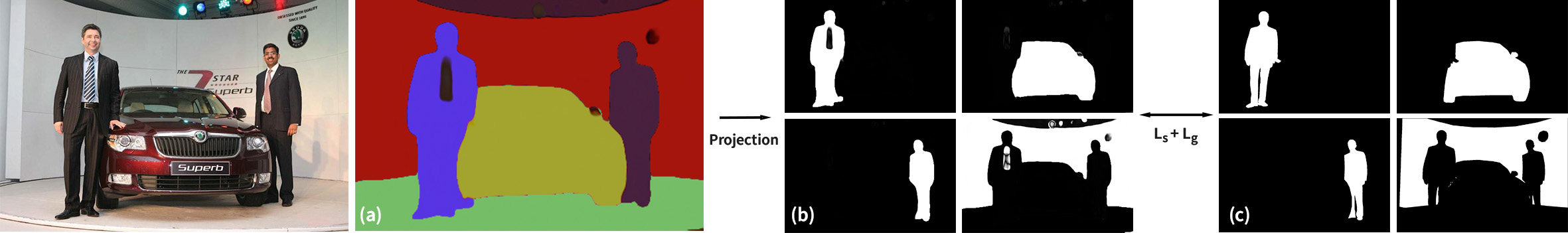}
  \caption{We show our resulting feature map in (a), reduced with PCA and colorized for visualization. Our projection into the segment space results in a set of binary maps in (b) that are compared against the ground-truth in (c) via our segmentation loss.}
  \label{fig:method_projection}
\end{figure}
\subsection{Segment-space Representation}
In order to directly evaluate the clustering performance of the generated features, we compute per-entity segmentation maps by defining a smooth linear transformation from our feature-space to what we call the \emph{segment-space}. 
Given $K$ known entities excluding the background, we define segment-space as the $(K-1)$-dimensional projective sphere, as represented by the unit sphere in $\mathbb{R}^K$ with $\vec{\mu}_k$ serving as basis vectors. 
We define a dimension-reducing linear transformation matrix $P$ from feature-space to segment-space such that segment $k$ is represented by the $k$th unit vector $\vec{e}_k$:
\begin{equation}
P \vec{\mu}_k = \vec{e}_k \quad \forall k
\end{equation}
which we compute as the left-inverse of the matrix $A = [\vec{\mu}_1 \quad \vec{\mu}_2 \quad \cdot\cdot\cdot \quad \vec{\mu}_K]$ 
so long as such an inverse exists. 
In the case that the network fails to differentiate between two entities during training, i.e. $\vec\mu_k \approx \vec\mu_l$ for any $k$ and $l$, the solution becomes degenerate. 
In such cases, we exclude our segmentation-space losses from back-propagation\footnote{These cases appear in less than $0.02\%$ of iterations during early stages of training and their frequency drops with further training.}.
By transforming the feature vectors for each pixel to this sub-space with $P\cdot\vec{f}_i,  \forall i$, we get a $h\times w$ map with $k$ channels where the absolute value of the $k$th channel is a real-valued map in $[0,1]$ that represents the alignment of each feature to $\vec{\mu}_k$.
We will denote each channel in our segment-space with $S_k$.
As the transformation $P$ makes all target lines orthogonal, the losses we define in the segment-space are amplified within the space between any not-yet-orthogonal $\vec\mu_k$ and $\vec\mu_l$.
This helps sort out features between entities that are not yet fully differentiated.
As all background features should be orthogonal to all entities, their features should lie squarely in the null-space of $P$ and their projections can be pushed towards $\vec{0}$. 

\begin{figure}[t]
  \centering
  \includegraphics[width=\linewidth]{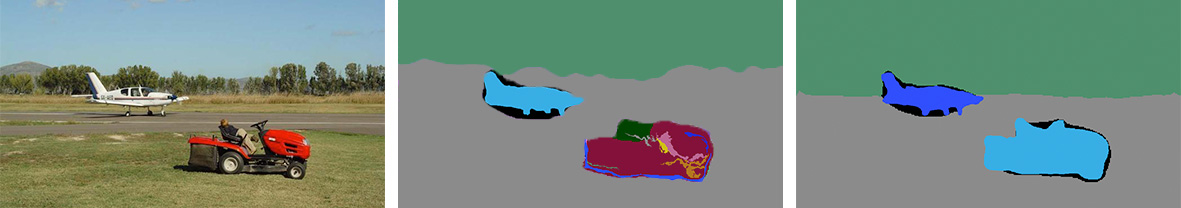}
  \caption{Two predictions from early training phases, (a) with only contrastive supervision and (b) with our segmentation loss added.}
  \label{fig:segmentation_loss}
\end{figure}

As we aim to estimate features $\vec{f}_i$ that are aligned to $\vec{\mu}_k$ for $i \in \mathcal{E}_k$ and orthogonal for $i \notin \mathcal{E}_k$, the ground-truth for $S_k$, which we denote as $\hat{S}_k$, is the binary ground-truth segmentation map for the $k$th entity.
We define our \emph{segmentation loss} as the mean-squared error over each channel
together with a gradient-based loss defined over multiple scales \cite{li2018megadepth} to enforce spatial smoothness:
\begin{equation}
    \mathcal{L}_s = \frac{1}{K} \sum_k MSE\left(S_k, \hat{S}_k\right), \quad\quad 
    \mathcal{L}_g = \frac{1}{K} \sum_k \sum_m MSE \left( \nabla S_k^m, \nabla \hat{S}_k^m \right),
\end{equation}
where $\nabla S_k^m$ is the spatial gradient of $S_k$ at the $m$th scale. We visualize the projection and segment supervision in \reffig{fig:method_projection}. 
Transforming our features into our segment-space representation allows us to use this standard multi-scale gradient loss defined on single channels for promoting spatial consistency in our hyper-dimensional features that lie in projective space by back-propagating this loss through the linear mapping $P$.
We show the effect of our segmentation loss in~\reffig{fig:segmentation_loss} on the outputs of two toy networks after $8$ epochs in training, one of which is trained exclusively on our contrastive losses, the other with the addition of the segmentation and multi-scale gradient losses. The contrastive losses enable the network to distinguish visual elements, while the additional segmentation losses enforce smooth features and turn the clusters into meaningful segments.

\subsection{Network Architecture and Training}
We define our final loss function as:
\begin{equation}
    \mathcal{L} = 
    \mathcal{L}_a + \mathcal{L}_r +  \mathcal{L}_s + 
    \lambda_{rc} \mathcal{L}_{rc} +
    \lambda_g \mathcal{L}_g + \lambda_{u} \mathcal{L}_{u},
\end{equation}
where $\lambda_{rc}$, $\lambda_g$, $\lambda_{u}$ are set to be 0.125, 0.025, and 0.05, respectively.
We employ the encoder-decoder architecture proposed by \cite{xian2018monocular} as our feature generator, but replace the encoder with the base ConvNext~\cite{liu2022convnet} as the backbone. Following Ranftl~\etal~\cite{midas}, we add four chained RefineNet~\cite{lin2019refinenet} modules as decoder block, followed by three convolutional layers to upscale the features back to the input size. A final \texttt{tanh} activation generates the output features.

\section{Inference-time Clustering}
\label{sec:inference}
Our network is designed to generate pixel features that are aligned together for pixels that belong to a single entity. Our training process involves utilizing repulsion and segmentation losses to ensure that features associated with a specific entity are orthogonal to those of other entities. This per-pixel representation effectively distinguishes between different objects, facilitating the application of a simple clustering method for segmentation.

In our approach, the classical mean-shift clustering method aligns seamlessly with our mean-based representation. Specifically, we apply mean-shift clustering on the $d$-dimensional hyper-sphere with a bandwidth set to $\frac{\sqrt{2}}{2}$, representing a 45\textdegree\ separation. This choice leverages the orthogonal inter-entity features to cluster the image into distinct segments.

\subsection{Multi-resolution Refinement}
\label{sec:multires}
In our proof-of-concept implementation, we leverage a simple convolutional neural network (CNN) for its effectiveness in achieving robust training, even with small datasets, rendering it well-suited for our application. The inherent limitation of CNNs, where their reasoning capability is confined to the size of their receptive field, manifests in coherent outcomes at this resolution. However, as demonstrated in other mid-level vision tasks~\cite{miangoleh2021boosting,careagaIntrinsic}, the capacity to generate intricate details significantly improves when inference is conducted at higher resolutions. This enhancement, however, comes at the cost of global coherency, as at higher resolutions, the network can only reason about small patches in the image at once, over-segmenting larger ones.

To capitalize on the consistency at smaller resolutions and the capacity to generate intricate segmentations at higher resolutions, we implement a multi-resolution clustering strategy. This involves feature generation and mean-shift clustering at multiple resolutions. The final segmentation map is constructed by processing segments from the smallest resolution first. With each increasing resolution, we incorporate clusters contained within existing segments in our map. New segments with a very high overlap with an old segment represent a refined version of the same segment with higher boundary accuracy, while segments that subdivide existing segments reveal smaller objects detected in the image. Our low-to-high resolution cluster merging strategy 
enables the generation of highly detailed segmentation maps in in-the-wild images with varying contexts. We provide a detailed description of our multi-resolution refinement approach in the supplementary material.

\section{Experiments}
\label{sec:experiments}
We train our model for 60 epochs with a learning rate of $1 \times 10^{-5}$. We use ImageNet~\cite{imagenet}-pretrained weights for the encoder and train all other modules from scratch.
We use the class-based COCO Panoptic~\cite{caesar2018coco}, ADE20K~\cite{zhou2018ade20k}, and CIHP~\cite{cihp} datasets as well as procedurally generates images of simple geometric shapes for training.
We give a detailed description of our training process and extend our experimental analysis in the supplementary material.

\subsubsection{Class-agnostic Baselines}
We evaluate our method against recent class-agnostic segmentation methods Open-World Entity Segmentation (OWES) \cite{qiOpenWorldEntitySegmentation2021}, High-Quality Entity Segmentation (HQES) \cite{qilu2022finegrained}, and the Segment Anything (SAM) model~\cite{kirillov2023segment} with the smaller ViT-B encoder as it has a similar number of parameters to ours.
Our method and OWES~\cite{qiOpenWorldEntitySegmentation2021} use common class-based datasets as real-world training data.
HQES~\cite{qilu2022finegrained} trains solely on their novel high-resolution class-agnostic dataset of 600K annotations in 33K images targeting high recall in-the-wild.
SAM~\cite{kirillov2023segment} trains their model on their novel large-scale dataset of 1B annotations in 11M images in addition to the common class-based datasets.

\subsubsection{Class-based Baselines}
We also include several panoptic segmentation methods in our analysis, namely OneFormer~\cite{jain2022oneformer} and Mask2Former~\cite{cheng2022masked} as task-unifying segmentation approaches, MaskDINO~\cite{maskdino} being optimized for object detection, and kMaX-DeepLab~\cite{yu2022k} inspired by k-means clustering. 
As class-based approaches do not allow for training with multiple datasets due to class definition conflicts, we use their models trained on ADE20k~\cite{zhou2018ade20k}.

\subsubsection{Metrics}
We compute Recall as a measure of completeness, treating every segment as a true positive that has an $\mathrm{IoU}>0.5$ with a ground truth segment. 
Recall being a critical measure of performance for class-agnostic evaluation, we also report the percentage relative improvement $\Delta\%$ of all methods with respect to the baseline with the lowest recall. 
As a measure of segmentation accuracy, we use the standard mask-based intersection-over-union (IoU) metric as well as the Boundary IoU metric \cite{cheng2021boundary} that focuses on boundary accuracy of the segments.
The class-agnostic segmentation methods including ours, OWES \cite{qiOpenWorldEntitySegmentation2021}, HQES\cite{qilu2022finegrained}, and SAM~\cite{kirillov2023segment}, will naturally generate segments that may not be included in the ground-truth maps, as it is very challenging to annotate every object in every scene. 
This makes the evaluation over false positives not a meaningful metric for this task, hence we exclude it.

\subsection{Evaluation over a Class-based Train/Test Split}
Class-agnostic segmentation aims to generalize to objects of any class, characteristically being trained on multiple class-based datasets or large class-agnostic datasets. 
Class-based segmentation, on the other hand, focuses on a specific set of classes in a fixed domain of images. 
In order to measure the effect of class-agnostic generalization on the performance on known classes, we present an evaluation in the domain of class-based segmentation methods using the test 

\begin{wraptable}{r}{0.60\linewidth}
\vspace{-2mm}
\caption{Evaluation on ADE20k \cite{zhou2018ade20k}}
\begin{adjustbox}{width=\linewidth}
{\renewcommand{\arraystretch}{1.2}}
\begin{tabular}{l|c|cccc}

Method                                                & Architecture-\#Params  & mIoU$\ua$       &B. IoU $\ua$     &Recall$\ua$   & $\Delta$\%$\ua$     \\ \hline   
kMaX-Deeplab~\cite{yu2022k}        & ConvNext-L - 244m        &0.366            &0.332            &0.376 & 0 \\  
MaskDINO~\cite{maskdino}            & Swin-L - 223m        &0.373            &0.340            &0.387 & 3  \\
Mask2Former~\cite{cheng2022masked}  &Swin-L - 216m        &0.413            &0.375            &0.430     & 14     \\
OneFormer~\cite{jain2022oneformer}    &Swin-L - 219m        &\udl{0.455} &\topscore{0.418} &\udl{0.484}  & \udl{29}  \\
\hline
OWES~\cite{qiOpenWorldEntitySegmentation2021}   &Swin-L - 208m        &0.400            &0.357            &0.420  & 12    \\

Ours - base res.& ConvNext-B - 101m      &0.380            &0.327        &0.391 &4 \\
\textbf{Ours}  & ConvNext-B - 101m        &\topscore{0.462} &\udl{0.407}      &\topscore{0.494} & \topscore{31}
\end{tabular}
\end{adjustbox}
\label{tab:ade20k}
\vspace{-5mm}
\end{wraptable}

\noindent
split of ADE20K~\cite{zhou2018ade20k} in Table~\ref{tab:ade20k}. 
All methods in Table~\ref{tab:ade20k} use the training split of ADE20K in their training.

Our method with the multi-resolution refinement performs on-par with the class-based OneFormer~\cite{jain2022oneformer} despite our smaller architecture, with a slight improvement in recall. 
This shows that our clustering-based bottom-up approach does not lead to a drop in performance on known classes. 
We see a drop in performance for the class-agnostic baseline OWES despite it having seen these classes during training, pointing to the mixed dataset training in top-down approaches having an adverse effect in domain-specific scenarios.

\subsection{Zero-shot Class-agnostic Evaluation}
Reflecting the in-the-wild generalization motivation of class-agnostic segmentation, we perform zero-shot evaluations on two recent class-agnostic datasets EntitySeg~\cite{qilu2022finegrained} and SA-1B~\cite{kirillov2023segment} that are characterized by their high number of class-agnostic annotations per image and high-resolution input images. As these datasets were used to train the models of HQES~\cite{qilu2022finegrained} and SAM~\cite{kirillov2023segment}, respectively, we exclude these methods from corresponding tables.

We present our evaluation on the first subset of the SA-1B dataset \cite{kirillov2023segment} in Table~\ref{tab:sa1b}. 
We use the first 1000 images in the dataset for general performance, and 
\begin{wraptable}{r}{0.60\linewidth}
\vspace{-5mm}
\caption{Zero-shot evaluation on SA-1B\cite{kirillov2023segment}}
\begin{adjustbox}{width=\linewidth}
{\renewcommand{\arraystretch}{1.2}}
\begin{tabular}{l|cccc|cccc}
\multirow{2}{*}{Method}                     & \multicolumn{4}{c |}{General performance}                   & \multicolumn{4}{c}{High object count}  \\
            & mIoU$\ua$           & B.IoU $\ua$        & Recall$\ua$      & $\Delta$\%$\ua$  & 
            mIoU$\ua$         &B.IoU $\ua$      &Recall$\ua$    & $\Delta$\%$\ua$          \\ \hline
Mask2Former~\cite{cheng2022masked}                      &0.318                &0.306            &0.326   & 0           &0.263              &0.285            &0.274 & 0  \\
OneFormer~\cite{jain2022oneformer}                    &0.335                &0.339            &0.342      & 5            &0.294              &0.357            &0.305  & 11  \\
kMaX-Deeplab~\cite{yu2022k}                              &0.337                &0.337            &0.344       & 6           &0.295              &0.338            &0.308  & 12  \\ 
MaskDINO~\cite{maskdino}                             &0.341                &0.340      &0.348      & 7          &0.316              &0.364            &0.329  & 20  \\ 
\hline
OWES~\cite{qiOpenWorldEntitySegmentation2021}    &0.356          &0.338            &0.370     & 14       &0.349        &0.380      &0.366  & 34  \\
HQES~\cite{qilu2022finegrained} (CF/Swin-L/217 m)   &\udl{0.398}          &\udl{0.418}           &\udl{0.411}      & \udl{26}      &\udl{0.384}        &\udl{0.448}      &\udl{0.401}  & \udl{46}  \\
\textbf{Ours}                               &\topscore{0.457}     &\topscore{0.424} &\topscore{0.500}     & \topscore{53}  &\topscore{0.480}   &\topscore{0.460} &\topscore{0.529}   & \topscore{93}    \\ 
\end{tabular}
\end{adjustbox}
\label{tab:sa1b}
\vspace{-5mm}
\end{wraptable}
\noindent
also create a different split of 500 images with the highest number of annotated objects in the set to measure the true positive rates in complex scenes. 
As Table~\ref{tab:sa1b} demonstrates, we significantly improve over class-agnostic approaches OWES and HQES, doubling the improvement in recall of second-best HQES with respect to the Mask2Former~\cite{cheng2022masked} baseline.
Our performance comes despite our smaller architecture as well as the high-resolution class-agnostic dataset collected to train HQES for in-the-wild generalization. 
This shows the effectiveness of our bottom-up approach in generating a class-agnostic understanding of objectness from class-based datasets.
We observe an expected drop in performance for class-based approaches, as class-agnostic datasets naturally contain many objects that are outside the list of classes for which these networks are trained. 

Table~\ref{tab:entity} presents our evaluation on the EntitySeg dataset~\cite{qilu2022finegrained}.
We perform 
\begin{wraptable}{r}{0.45\linewidth}
\vspace{-5mm}
\caption{Zero-shot on EntitySeg \cite{qilu2022finegrained}}
\begin{adjustbox}{width=\linewidth}
{\renewcommand{\arraystretch}{1.2}}
\begin{tabular}{l|cccc}
Method                                                 & mIoU$\ua$       &B. IoU $\ua$     &Recall$\ua$  & $\Delta$\%$\ua$   \\ \hline     
OneFormer~\cite{jain2022oneformer}                  &0.409 &0.382   &0.423    &  0 \\
Mask2Former~\cite{cheng2022masked}                    &0.452 &0.403 &0.474     &   12  \\
MaskDINO~\cite{maskdino}                           &0.463 &0.423    &0.482   & 14 \\
kMaX-Deeplab~\cite{yu2022k}                    &0.488 &0.455 &0.524   &  24 \\\hline
OWES~\cite{qiOpenWorldEntitySegmentation2021}   &0.521 &0.470 &0.566     &   34  \\
SAM~\cite{kirillov2023segment}             &\topscore{0.585}  &\topscore{0.539} &\topscore{0.619}   &  \topscore{46} \\
\textbf{Ours}     &\udl{0.574}  &\udl{0.522} &\udl{0.614}  & \udl{45}
\end{tabular}
\end{adjustbox}
\label{tab:entity}
\vspace{-5mm}
\end{wraptable}

\noindent
on-par with SAM~\cite{kirillov2023segment} despite their use of a class-agnostic dataset that is larger than our training set by two magnitudes. 
This demonstrates that our carefully designed clustering-based approach is highly effective in generalizing to in-the-wild images without the need for a large-scale dataset.

Earlier attempts at class-agnostic segmentation \cite{hwang2021exemplar,xu2022dual,qiOpenWorldEntitySegmentation2021} focused on developing classless formulations and trained on available class-based datasets. 
This focus shifted to the collection of high-quality and large-scale datasets in recent literature~\cite{qilu2022finegrained,kirillov2023segment} due to the straightforward effectiveness of a well-crafted training dataset in generalization despite their cost. 
Our methodology stands orthogonal to the recent literature, where we achieve state-of-the-art performance in in-the-wild class-agnostic segmentation using a novel bottom-up approach to the segmentation problem using limited training data and a simple network architecture. 
While this divergence creates new promising research and development directions for class-agnostic segmentation, combining our clustering-based formulation with large-scale training procedures, 
it also enables the application of our methodology in other, data-scarce segmentation problems as discussed in the next section. 

\subsection{Evaluation on Cell and Nucleus Segmentation}
\label{sec:evican}
Our bottom-up approach is designed with a focus on simply generating orthogonal features for differentiating segments in an image. 
Our formulation
\begin{wraptable}{r}{0.75\linewidth}
\vspace{-5mm}
\caption{Evaluation on EVICAN dataset~\cite{schwendy2020evican}}
\begin{adjustbox}{width=\linewidth}
{\renewcommand{\arraystretch}{1.2}}

\begin{tabular}{l|ccc|ccc|ccc}
\multirow{2}{*}{Method}& \multicolumn{3}{c|}{Easy Difficulty} & \multicolumn{3}{c|}{Medium Difficulty} & \multicolumn{3}{c}{Hard Difficulty}    \\
                            & mAP$\ua$ & AP@50$\ua$ &AP@75$\ua$ & mAP$\ua$ & AP@50$\ua$ &AP@75$\ua$ & mAP$\ua$ & AP@50$\ua$ &AP@75$\ua$   \\ \hline
MRCNN~\cite{schwendy2020evican}    &0.322    &0.616 &0.317    &0.136    &0.310    &0.105      &0.085    &0.208     &0.044  \\
DeepCeNS~\cite{khalid2021deepcens}    &\topscore{0.526}     &\topscore{0.834}    &\topscore{0.573}            &\udl{0.261}    &\udl{0.479}     &\topscore{0.289}     &\udl{0.169}     &\udl{0.338}    &\udl{0.158}   \\
\textbf{Ours}                &\udl{0.408} &\udl{0.663}      &\udl{0.400}     &\topscore{0.304}      &\topscore{0.565}      &\udl{0.223}   &\topscore{0.322}      &\topscore{0.589}       &\topscore{0.290}
\end{tabular}
\end{adjustbox}
\label{tab:evican}
\vspace{-5mm}
\end{wraptable}
\noindent
being developed towards a simple clustering of features achieves state-of-the-art performance even when trained on a smaller training dataset. 
This makes our method directly applicable to segmentation problems in other domains such as biomedical images. 
We demonstrate this by training our network, without any changes in the formulation, on the EVICAN dataset~\cite{schwendy2020evican} for the problem of cell and nucleus segmentation. 
EVICAN dataset provides 4464 annotated microscopic images in their training split, where images are typically of very little contrast. 
Using their test set, divided by them into 3 difficulty levels, we compare against domain-specific biomedical segmentation approaches MRCNN~\cite{schwendy2020evican} and DeepCeNS~\cite{khalid2021deepcens} in Table~\ref{tab:evican}. 
Following Schwendy~\etal~\cite{schwendy2020evican}, we use the average precision (AP) with different IoU thresholds. 
We report the AP at 50 and 75\% IoU, as well as the mean value of AP's at 10 different IoU thresolds between 50-95\%.
We show competitive performance in easy and medium difficulties, while significantly improving upon the baselines in the hard difficulty subset. 
This demonstrates the generic nature of our bottom-up approach to segmentation.

The network architectures adopted by these domain-specific approaches are both convolutional neural networks similar to ours. 
Although CNN's have inherent limitations coming from a fixed receptive field, their ease of training makes them applicable to data-scarce domains. 
This is in contrast with transformer-based architectures such as HQES~\cite{qilu2022finegrained} and SAM~\cite{kirillov2023segment} that can not be trained using such small datasets. 
While our formulation can be applied to transformer-based architectures, we show that the receptive field limitation of CNN's can be remedied within our feature-based formulation with a simple multi-resolution estimation procedure as detailed in Section~\ref{sec:multires}.

\subsection{The Effect of the Segment-space Representation}
Our system makes use of metric-learning inspired attraction and repulsion losses $\mathcal{L}_a$ and $\mathcal{L}_c$ that allows the network to differentiate between entities. 
We also include the regional contrast loss $\mathcal{L}_{rc}$ for cases where the network groups two entities together, as detailed in Section~\ref{sec:metriclosses}. 
In order to improve clustering quality, we develop our segment-space representation and define an MSE and a gradient-based loss,  $\mathcal{L}_s$ and $\mathcal{L}_g$, respectively. 
In this section, we measure the effect of our segment-space representation on our generated features. 
For this purpose, we measure the two qualities we expect from the generated features: inter-mean similarity and intra-entity similarity. 
Inter-mean similarity measures if the network is able to generate different mean orientations $\vec{\mu}_k$ for different entities, computed as the average cosine similarity between each $(\vec{\mu}_{k}, \vec{\mu}_l)$ pair in an image.
\begin{wraptable}{r}{0.65\linewidth}
\vspace{-5mm}
\caption{We compute the cosine-similarity between means of different entities as well as features within and entity for different loss combinations on the ADE20K~\cite{zhou2018ade20k} dataset.}
\begin{adjustbox}{width=\linewidth}
{\renewcommand{\arraystretch}{1.2}}
\begin{tabular}{l|cc}
Included losses                        & Inter-mean sim.$\da$ (deg. $\ua$)    & Intra-entity sim.$\ua$ (deg. $\da$)
\\ \hline     
$\mathcal{L}_{rc}$ +  $\mathcal{L}_a$ + $\mathcal{L}_r$               
                        & \topscore{0.024} \quad (88.6\textdegree)      &0.152 \quad (81.2\textdegree)
                        \\
$\mathcal{L}_{rc}$ +  $\mathcal{L}_a$ + $\mathcal{L}_r$ + $\mathcal{L}_s$                        
                        &0.031   \quad (88.2\textdegree)    &\udl{0.745}  \quad (41.8\textdegree)
                        \\
$\mathcal{L}_{rc}$ +  $\mathcal{L}_a$ + $\mathcal{L}_r$ + $\mathcal{L}_s$ + $\mathcal{L}_g$
                       &\udl{0.029} \quad (88.3\textdegree)          & \topscore{0.754} \quad (41.1\textdegree)
                       \\ 
\end{tabular}
\end{adjustbox}
\label{tab:ablations}
\vspace{-5mm}
\end{wraptable}

\noindent
If the network is able to generate perfectly orthogonal orientations for every entity in the image, our inter-mean similarity metric would be 0.
Intra-entity similarity is computed as the average cosine similarity between the mean assigned to each entity and the features generated for pixels belonging to that entity.
It measures how well-aligned the features belonging to an entity to their corresponding mean orientation $\vec{\mu}_k$.
In the case perfect alignment, our intra-entity similarity metric would be 1.
A good performance in both metrics is desired for effective clustering of our features.

We compare the improvement brought by $\mathcal{L}_s$ and $\mathcal{L}_g$ in Table~\ref{tab:ablations}. 
We perform this test by training 3 different versions of our formulation on the ADE20k training set for 10 epochs. 
It should be noted that all 3 versions include our differentiation-based losses.
This is required for our network to generate different features for different entities, as our segment-space based losses only promote better alignment inside a given entity.
As Table~\ref{tab:ablations} shows, our discriminative losses are effective in generating almost perfectly orthogonal orientations for different entities in all scenarios, while struggling to generate well-aligned features for each segment.
Our segment-space loss $\mathcal{L}_s$ significantly improves the performance in this aspect, driving intra-entity similarity below the critical 45\textdegree threshold that represents clear separation between the features of two entities with orthogonal orientations. 
Our gradient-based loss $\mathcal{L}_g$ further improves the performance both in terms of inter-mean orthogonality and intra-entity alignment.

\section{Conclusion and Limitations}
In this paper, we introduced a bottom-up approach to class-agnostic segmentation and show that this generalist approach shows a great generalization ability to out-of-distribution images despite using standard class-based segmentation datasets as real-world training data. 
This is in line with our current understanding of human cognition which is modeled as a bottom-up process.
Our network is trained with a novel formulation that integrates ideas from contrastive learning with our new segment-space representation to achieve detailed clustering of semantically meaningful regions in any image. 
We demonstrate the performance of our formulation using a proof-of-concept implementation with a small convolutional architecture trained on around 200k real-world images with class-based annotations.
Our approach represents a promising new direction for in-the-wild class-agnostic segmentation that is typically approached with top-down formulations as well as diverse segmentation challenges in other domains with limited training data.
We make use of iterative clustering of objects on image patches to generate high-resolution segmentation results. 
This circumvents the limited receptive field size of convolutional neural networks. 
The performance of our method can be further improved by utilizing network architectures with higher number of parameters and higher receptive field size.

We so far have only cared for entities to be different, using the segments included in class-based datasets as individual entities. 
However, this entity definition does not reflect the complexity of the real world. 
The definition of an entity varies depending on the context. 
For instance, while the common datasets treat \emph{person} as a single entity, there are many applications that would benefit from the segmentation of body parts or facial features. 
We believe a hierarchical representation of entities will allow class-agnostic segmentation to be applicable to a wider range of problems.

\section*{Acknowledgements}
We acknowledge the support of the Natural Sciences and Engineering Research Council of Canada (NSERC), [RGPIN-2020-05375].

{\small
\bibliographystyle{splncs04}
\bibliography{segmentation_full}
}

\end{document}